\documentclass[letterpaper]{article} 
\usepackage[]{aaai2027}  
\usepackage[hyphens]{url}  
\usepackage{graphicx} 
\usepackage{natbib}  
\usepackage{caption} 
\usepackage{algorithm}
\usepackage{algorithmic}
\usepackage{booktabs}
\usepackage{array}
\usepackage{amssymb} 
\usepackage{amsmath}
\usepackage{multirow}
\usepackage{makecell}
\usepackage{newfloat}
\usepackage{listings}
\DeclareCaptionStyle{ruled}{labelfont=normalfont,labelsep=colon,strut=off} 
\floatstyle{ruled}
\newfloat{listing}{tb}{lst}{}
\floatname{listing}{Listing}

\usepackage{booktabs}
\usepackage{subcaption}

\title{SkyDrive: Learning to Drive in a New City from Aerial Traffic Monitoring}
\author{
    Weijiang Xiong,
    Lan Feng,
    Alexandre Alahi,
    Nikolas Geroliminis~\corresponding
}
\affiliations{
    \textsuperscript{\rm 1}École Polytechnique Fédérale de Lausanne\\

    firstname.lastname@epfl.ch
}

\begin{document}

\maketitle

\begin{abstract}
Autonomous driving has made remarkable progress through imitation learning with massive human demonstration data.
However, a trained planner often degrades severely when applied to a new environment zero-shot, because of domain shifts in traffic regulations, road layout and driving behaviors.
Therefore, adapting a trajectory planner to a new city typically requires resource-demanding local data collection with a vehicle sensor suite.
In this work, we show that driving behavior can be learned from a scalable and efficient alternative.
We introduce \emph{SkyDrive}, a framework that utilizes drone-based traffic monitoring to provide efficient supervision for autonomous driving agents in a new environment.
While vehicle-based data collection logs the ego and its surroundings, an aerial platform naturally observes many road users simultaneously over an extended field of view.
As a result, every vehicle can be a data source with grounded driving behavior, effectively scaling up the amount of supervision.
Based on 137 hours of aerial traffic monitoring footage, we extract 650K driving samples and construct a benchmark for trajectory planners and motion predictors.
Zero-shot experiments with multiple models reveal significant cross-city domain gaps, but many of them can be alleviated by limited supervision from the sky, e.g., 30 minutes of monitoring per location.
Our findings show that aerial traffic monitoring is an efficient and scalable data source for adapting autonomous driving systems in new cities.
Data and code will be made publicly available.

\end{abstract}


\section{Introduction}
\label{sec: introduction}
Learning to imitate human driving behavior has been the cornerstone of autonomous driving.
Trained on sufficient and high-quality expert demonstrations, such as Argoverse, nuPlan, and the Waymo Open Dataset, modern trajectory planners can learn to navigate a vehicle under complex scenarios \citep{benjamin2021argoverse2,caesar2021nuplan,ettinger2021waymodataset}.
However, the strong in-domain capability often does not guarantee reliable transfer to an unseen environment, e.g., a new city, due to the recognized out-of-distribution problem \cite{zhou2023domain}.
Since geographical distribution shifts can arise from various aspects, including landscape, road layout and traffic regulations, performance degradations are often expected in object detection \citep{chen2018domain, yang2021st3d}, motion prediction \citep{yao2024improving} and trajectory planning \citep{yasarla2025roca}.
Therefore, zero-shot transfer is prone to pronounced errors \cite{naeinian2026zero}.

To tackle this challenge, a direct solution is to retrain the model after collecting and annotating enough driving logs with instrumented vehicles in the target city.
But conducting such an ambitious project for every new city demands lots of resources, especially for covering the long-tailed distribution of driving scenarios.
Fortunately, in addition to the ego vehicle, surrounding agents can also provide valuable planning supervision \cite{chen2022lav}, which greatly improves data efficiency. 
However, unclear driving intentions, fragmented trajectories and perception noise have become the major obstacles due to sensor range limitations and occlusions \cite{zhang2021learning}.
Inspired by this, we opt for an alternative solution with minimal occlusion, clear visibility and an extended field of view, i.e., aerial traffic monitoring \citep{fonod2025advanced}.
Such a platform can simultaneously detect and track many vehicles within the view, and all visible vehicles can be turned into driving demonstrations, making it an efficient and scalable data source.

We introduce \emph{SkyDrive}, a framework that constructs driving scenarios from aerial traffic monitoring.
Building upon geo-referenced vehicle tracks, \emph{SkyDrive} carefully selects agents as virtual egos and extracts driving scenes centered on them.
The scenes are rasterized into ego-centric multi-view semantic images for trajectory planning, and exported as vectorized tracks for motion prediction, allowing diverse supervision grounded on real driving behaviors.
The benchmark reveals pronounced zero-shot degradation under domain shifts and shows that a small subset of monitoring sessions can substantially improve target-domain performance.

In summary, this work makes the following contributions:
\begin{itemize}
    \item A scalable and efficient pipeline that converts aerial traffic monitoring into localized driving supervision.
    \item A high-quality dataset with $\sim$650K real scenarios from 137.2 hours of aerial observations over 20 complex modern intersections.
    \item Detailed experiments on trajectory planning and motion prediction to demonstrate the zero-shot transfer challenge and the efficiency of aerial supervision.
\end{itemize}

\section{Related Work}
\label{sec: related work}




\subsection{Autonomous Driving}
\label{sec: autonomous driving}

Motion prediction is an important block in earlier paradigms of autonomous driving with modularized perception, prediction and planning.
AutoBot jointly encodes the motion of all agents and decodes map-consistent futures \citep{girgis2022latent}.
MTR designs learnable queries to represent different driving intentions \citep{shi2022mtr}.
Wayformer applies early fusion for heterogeneous information, including traffic light, map, ego history and surrounding agents~\citep{nayakanti2023wayformer}.
UniTraj evaluates them on multiple datasets and reveals significant cross-domain gaps~\citep{feng2024unitraj}.

Since modularized systems may be suboptimal for the ultimate goal due to accumulated errors \citep{chen2023e2esurvey}, most modern solutions train end-to-end models to learn trajectory planning directly from sensor observations.
For example, TransFuser combines camera and LiDAR features for waypoint prediction \citep{prakash2021transfuser}.
The Bird's-Eye-View (BEV) is a central latent space for fusing multi-modal sensor inputs and learning various representations.
UniAD pivots to trajectory planning and learns perception and prediction as parallel tasks \citep{hu2023uniad}.
VAD decodes the agent motion and map structure as vectors \citep{jiang2023vad}.
FlowDrive predicts an interpretable flow field for safer planning \citep{jiang2025flowdrive}.
Since the BEV space can be heavy, recent works have proposed to generate trajectory plans directly from multi-view camera features, e.g., SparseDrive \citep{sun2025sparsedrive} and DrivoR \citep{kirby2026drivor}.

Driving in the real world involves considerable uncertainty, and thus modern planners often generate and score multimodal predictions.
VAD-V2 builds a planning vocabulary with possible trajectories, and predicts a probability distribution over the action space \citep{chen2024vadv2}.
Hydra-MDP learns to score the vocabulary with both human demonstration and rule-based experts \citep{li2025hydramdp}.
DiffusionDrive \citep{liao2025diffusiondrive} denoises anchored Gaussian distributions, and GoalFlow \citep{xing2025goalflow} generates a goal-conditioned plan with Flow Matching.

\subsection{Synthetic Views in Autonomous Driving}
\label{subsec: synthetic views in autonomous driving}

Traffic simulators such as CARLA, MetaDrive and Hugsim are important platforms for training and evaluation~\citep{dosovitskiy2017carla,li2022metadrive, zhou2024hugsim}.
From simulation, diverse driving scenarios and rendered sensor observations can be generated, which are essential for safety-critical decisions \citep{liu2026advbmt}.
With the recent advancements in neural rendering, such as Gaussian Splatting, more realistic 3D scenes can be built from the real-world driving logs \citep{kerbl3Dgaussians}.
SimScale samples trajectories from reconstructed 3D scenes and uses them as augmented data for training more robust policies \citep{tian2025simscale}.
RAD trains the driving policy to fully explore the scene via reinforcement learning (RL) \citep{gao2026rad}.

Another thread of work has focused on the semantic layout of the scene instead of pursuing photorealism.
\citet{xu2018autonomous} and \citet{muller2018driving} train a driving policy based on semantic segmentation results of the ego view.
\citet{chung2022segmented} and \citet{behl2020label} propose to abstract away the visual appearance details with instance-level segmentation masks.
RAP renders the camera views for the ego agent using 3D bounding boxes of traffic participants, and aligns the feature space distribution of real camera input with the rendered semantic views.
As a result, the model can focus on the semantics of the scenes, e.g., lane centerlines, road boundary and surrounding agents, and stay robust to the drifts in visual appearance.
Gigapixel similarly renders a simplified bounding-box world to support pixel-space RL-based self-play~\citep{rowe2026gigapixel}.
Such semantic layouts have also been utilized as control images in Cosmos3 for rendering photo-realistic views \citep{agarwal2026cosmos}.

\subsection{Aerial Traffic Monitoring}
\label{subsec: aerial traffic monitoring}
Drones are popular platforms for collecting trajectory data due to their high mobility and the low occlusion of aerial perspective.
For example, HighD and inD record traffic on highways and at intersections, respectively~\citep{krajewski2018highd,bock2020ind}.
SinD additionally includes traffic signal states \citep{xu2022sind}, Interaction emphasizes the joint behavior of vehicles \citep{zhan2019interaction}, and HeteroD points out the importance of vulnerable road users~\citep{chen2026hetrod}.
However, these datasets are small-scale (e.g., a single drone and a few hours) and are often scattered across many cities.
Meanwhile, the transportation community has initiated large-scale coordinated monitoring with multiple drones over continuous urban regions.
The pNEUMA experiment monitors the entire city center of Athens, Greece \citep{barmpounakis2020pneuma}, and Songdo Traffic contains geo-referenced vehicle trajectories for complex intersections in Songdo, South Korea \citep{fonod2025advanced}.
SkyDrive repurposes the Songdo Traffic data from mobility analysis to autonomous driving by treating each tracked car as an ego vehicle and building egocentric scenarios.

\section{SkyDrive: Learn to Drive from the Sky}
\label{sec: skydrive learn to drive from the sky}

\subsection{Overview}
\label{subsec: overview}
\begin{figure*}[ht]
    \centering
    \includegraphics[width=\textwidth]{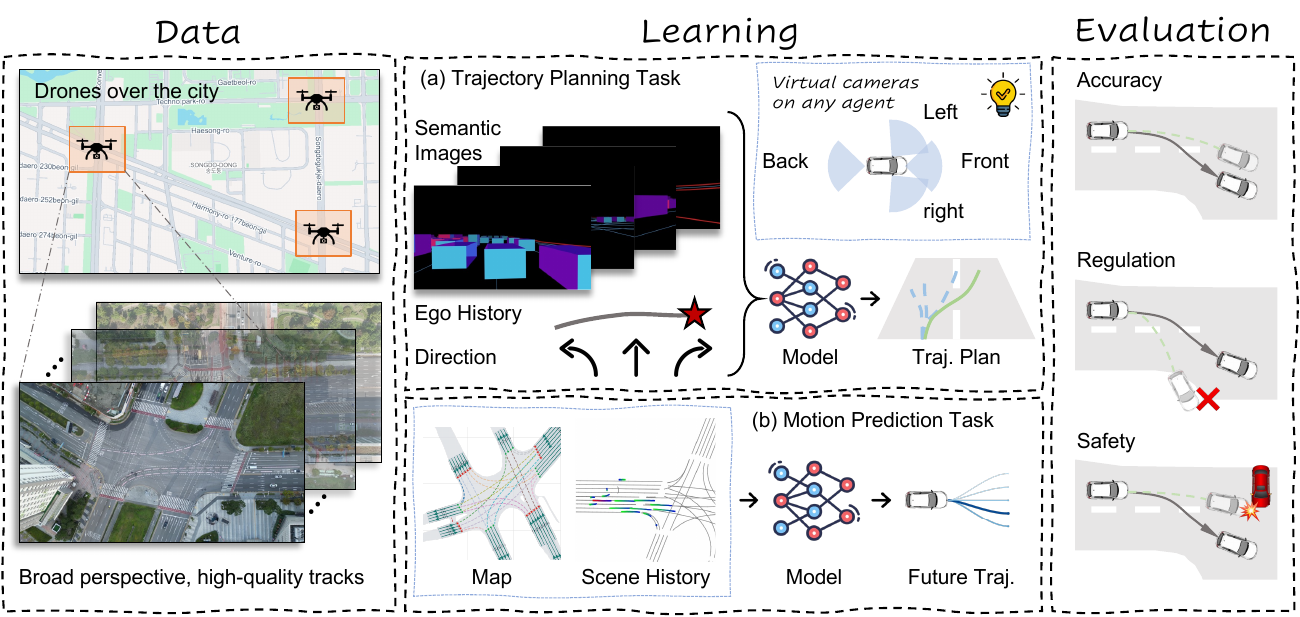}
    \caption{Overview of \emph{SkyDrive}. We utilize accurate vehicle trajectory data from drone-based traffic monitoring research, study trajectory planning and motion prediction tasks, and evaluate accuracy, regulatory compliance and safety.}
    \label{fig: workflow}
\end{figure*}

Figure~\ref{fig: workflow} presents the overall workflow of \emph{SkyDrive}, which turns aerial traffic monitoring into real driving behavior supervision.
In the data collection stage, a swarm of drones is deployed over the city to monitor multiple locations simultaneously.
Then, the accurate locations of most vehicles can be continuously tracked for the entire duration of their presence in the bird's-eye-view videos, encoding the driving behavior of those traffic participants.
Therefore, the egocentric dynamics of any agent can be learned by centering the coordinate system on it, which brings two-fold benefits. 
First, the vehicle driving behaviors can be learned without operating instrumented vehicles for a long time.
Second, the data volume can be effectively scaled up while ensuring the quality.

With the trajectory data and map information, we primarily focus on vision-based trajectory planning where a model plans the future ego trajectory using semantic images, ego history and a high-level driving direction.
We evaluate the planners in terms of accuracy, regulation compliance and safety.
Meanwhile, we also experiment with motion prediction where a model predicts possible future ego motions with the map and historical trajectories.
The remainder of this section will describe the dataset and the tasks in detail.

\subsection{The SongdoDrive Dataset}
\label{subsec: the songdodrive dataset}

We build our dataset with the Songdo Traffic \citep{fonod2025advanced} dataset, a traffic monitoring dataset collected by a swarm of drones from the modern city of Songdo in South Korea.
The dataset provides high-quality geo-referenced vehicle coordinates along with annotated lane bounding boxes, and has been widely utilized in traffic analysis problems.
In this work, we repurpose it for trajectory planning as well as motion prediction.
To this end, we have developed a data processing pipeline to extract driving scenes, and we refer to the derived dataset as SongdoDrive.

In Songdo Traffic \citep{fonod2025advanced}, the trajectory data is collected from 20 complex urban intersections (e.g., labeled as A, B or C).
The experiment spans four days and each day has 10 monitoring time slots from morning to evening (e.g., 8:00 -- 8:30).
Although the virtual cameras can be placed on any vehicle, we focus on the cars with valid movements, i.e., the 50\% speed quantile is above 10 km/h.
Then, we progressively select instances from the candidate pool, and sample 8-second segments with a stride of 4 seconds. 
The segments with insufficient length, missing locations or invalid size estimations are discarded.
In addition, near-duplicate segments are filtered out, e.g., two cars closely following each other.
These selected segments are regarded as ego movements, and a driving scenario is then sliced from the monitoring session, which means a scenario contains the complete view of the intersection during the time span of the ego segment.

\begin{figure}[ht]
    \centering
    \begin{subfigure}{0.48\linewidth}
        \centering
        \includegraphics[
            height=\linewidth,
            trim=5cm 5cm 5cm 5cm,
            clip
        ]{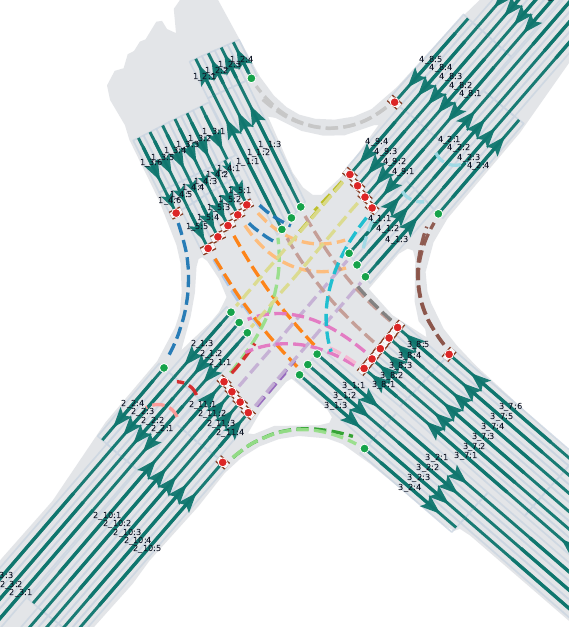}
        \caption{B}
    \end{subfigure}
    \hfill
    \begin{subfigure}{0.48\linewidth}
        \centering
        \includegraphics[
            height=\linewidth,
            trim=5cm 5cm 5cm 5cm,
            clip
        ]{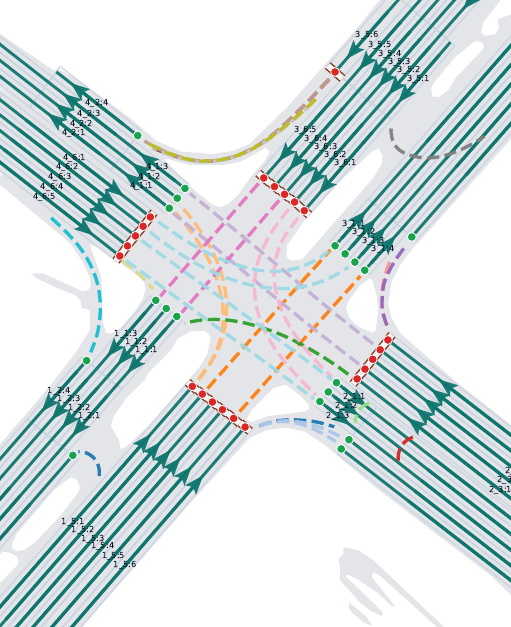}
        \caption{K}
    \end{subfigure}

    \caption{Map visualizations for two example intersections. The lane bounding boxes are labeled in Songdo Traffic, and the stop lines, lane centerlines, and drivable area are derived using trajectory data.}
    \label{fig: maps of intersections}
\end{figure}

Using the trajectories and lane bounding boxes, we also derived various types of map information, including the lane directions, stop lines, drivable area polygons and lane connectors.
Two such examples are shown in Figure~\ref{fig: maps of intersections}, and more of those preprocessing details are in the supplementary materials.

After the preprocessing, SongdoDrive has 650K scenarios from 137.2 hours of drone monitoring logs, and Table~\ref{tab: statistics of standard data split} shows the overall statistics and the standard train-test splits.
After the initial near-static speed filtering, about 368K car trajectories remain, and then $\sim$234K segments are rejected in subsequent data quality control.

\begin{table}[ht]
    \centering
    \caption{Statistics of the standard data split}
    \label{tab: statistics of standard data split}
    \resizebox{\columnwidth}{!}{%
    \begin{tabular}{@{}lcrrrr@{}}
        \toprule
        {Split} & {Sessions} & {Time (h)} & {Valid veh.} & {Filtered} & {Ego segs.} \\
        \midrule
        Train & 680 & 116.23 & 313075 & 200440 & 549929 \\
        Test & 120 & 20.97 & 54731 & 33910 & 99629 \\
        Overall & 800 & 137.20 & 367806 & 234350 & 649558 \\
        \bottomrule
    \end{tabular}
    }
\end{table}

Table~\ref{tab: comparison with other datasets} compares SongdoDrive with other motion datasets.
NuScenes, Argoverse2 and Waymo are collected with vehicle platforms, while the others are based on drones.
The trajectories from drone platforms are notably longer than those from vehicles, and therefore, more ego segments can be sampled from one trajectory.
As a result, SongdoDrive has a number of scenarios comparable to those of Waymo and Argoverse2 while requiring less data collection time.
Meanwhile, SongdoDrive is larger in scale than other drone-based datasets, providing more comprehensive coverage of driving behaviors.
Besides, the drones in SongdoDrive are deployed over the same urban region, making it a focused solution for adapting autonomous driving models to a specific city.

\begin{table}[ht]
\centering
\caption{Comparison with other datasets.}
\label{tab: comparison with other datasets}
\resizebox{\columnwidth}{!}{%
\begin{tabular}{@{}lccccc@{}}
\toprule
\multirow{2}{*}{Dataset} & \makecell{\# unique\\tracks} & \makecell{Avg track\\length} & \makecell{\#sce.} & \makecell{Scenario\\duration} & \makecell{Total\\time} \\
\midrule
NuScenes & 4.3k & -- & 50K & 8s & 5.5h \\
Argoverse2 & 13.9m & 5.16s & 250K & 11s & 763h \\
Waymo & 7.64m & 7.04s & 576K & 9.1s & 574h \\
\midrule
Interaction & 40k & 19.8s & -- & -- & 16.5h \\
SinD & 13.2k & -- & -- & -- & 7.02h \\
HeteroD & 64.5k & -- & -- & -- & 17.5h \\
\midrule
SongdoDrive & 646.5k & 33.8s & 650k & 8s & 137.2h \\
\bottomrule
\end{tabular}%
}
\end{table}

\subsection{Trajectory Planning Task}
\label{subsec: trajectory planning task}

\begin{figure*}[!t]
    \centering
    \includegraphics[width=\textwidth]{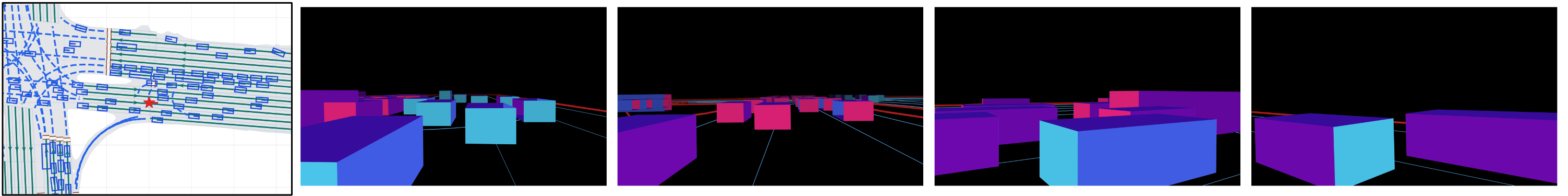}
    \caption{Example of semantic views during a lane-changing event in busy traffic. The leftmost panel shows a bird's-eye view of the scene, where the ego is rendered as a red star. The perspective views from left to right are front, back, left and right.}
    \label{fig: rendering example}
\end{figure*}

Following Navsim \citep{dauner2024navsim}, a trajectory planner uses the first 2 seconds of an 8-second ego segment as input information, and is required to plan the ego trajectory for the next 4s.
Both the history and prediction are required at 2Hz, with data and ground truth down-sampled from original 30Hz data.
Thus, the model has 4 input frames (including the current one) and predicts the next 8 steps.
Since the trajectory data is two-dimensional, we lift the driving scene into 3D space by assuming a flat ground surface and a common-sense per-type vehicle height.
Then, a set of virtual cameras is applied to the ego vehicles as shown in Figure~\ref{fig: workflow}, and the egocentric views can be rendered by rasterizing the 3D vehicle boxes and the map structures.
This rendering process follows RAP \citep{feng2026rap}, with camera matrices borrowed from \citep{caesar2021nuplan}. 
Figure~\ref{fig: rendering example} shows an example of rendered images in busy traffic.

In addition to the rendered semantic images (2s history at 2Hz), the model also has access to precise ego history, including location, velocity and heading with respect to the current vehicle pose.
Considering an autonomous vehicle should know its navigation target, the short-term trajectory planner also knows a high-level driving direction, i.e., left, right or straight, which is obtained offline from the future trajectories.
Depending on the design, a planner may choose not to use historical information and proceed only with the current frame \citep{kirby2026drivor}.

To evaluate the planner, we use the following metrics to cover accuracy, regulation and safety aspects.
\begin{itemize}
    \item Average Displacement Error (ADE) is the average distance (in meters) between the planned trajectory and the ground truth over all the future steps. 
    \item Final Displacement Error (FDE) is the distance (in meters) between the planned waypoint and the ground-truth waypoint at the last prediction time step.
    \item Non-Compliant Trajectory (NCT) measures compliance with respect to the drivable area. We report the percentage of trajectories with any waypoints outside the drivable area polygon, i.e., driving off-road.
    \item Time-to-collision (TTC) infraction rate considers the driving risks. At each planned waypoint, we project the ego at a constant velocity and heading 0.3, 0.6 and 0.9 seconds ahead, and check if it collides with any other vehicles. Similar to Navsim \citep{dauner2024navsim}, the background vehicles will not react to the ego and will proceed according to the logged data. The infraction rate is the percentage of trajectories where the ego has TTC$\leq$0.9s at any waypoint.
\end{itemize}

Since straight moves account for the vast majority of real-world driving, the overall result is likely to be dominated by them.
Therefore, we report finer-grained metrics by trajectory types and by Kalman Difficulty \citep{feng2024unitraj} to evaluate the performance under typical driving scenarios.

\begin{figure}[h]
    \centering
    \includegraphics[width=0.8\columnwidth]{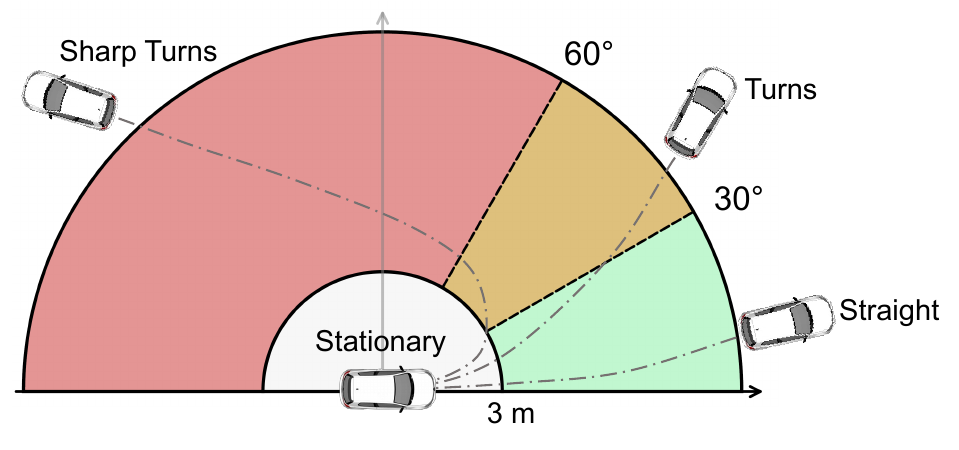}
    \caption{The trajectory types in the trajectory planning task}
    \label{fig: turning types}
\end{figure}

The trajectory types are decided according to the endpoint position (4s in the future), as illustrated in Figure~\ref{fig: turning types}.
All trajectories with displacement less than 3 meters are stationary.
Non-stationary trajectories are divided into straight trajectories, (normal) turns and sharp turns according to the angle between the initial heading and the endpoint displacement vector.
The turns have left and right directions, with a positive angle indicating left.
For example, a trajectory with endpoint (5m, +75\textdegree) from the start is a sharp left turn.
Unlike UniTraj, we do not distinguish U-turns since they can rarely be completed within the shorter prediction horizon.

Kalman Difficulty indicates how much the trajectory differs from naive movements.
A constant-velocity Kalman filter is estimated from past trajectories and projected to future steps.
The FDE between the Kalman filter prediction and the ground truth is then defined as Kalman Difficulty.
The trajectories are then grouped into easy, medium and hard categories based on Kalman Difficulty: $< 10$, $[10, 20)$ and $\geq 20$, respectively.

\subsection{Motion Prediction Task}
\label{subsec: motion prediction task}
The motion forecasting task follows UniTraj \citep{feng2024unitraj}.
A model has access to map structures, including lanes, stop lines and the drivable area, as well as past trajectories for the ego and surrounding vehicles.
Then, the model is required to predict $K=6$ possible future trajectories of the ego vehicle, along with the probabilities.
Concretely, the model receives 2s of history in each 8-second driving scenario and predicts 6s into the future, both at 10Hz.
Thus, the input contains 21 frames in total including the current time, and the output requires 60 future steps.
Besides, the coordinate system is centered on the current ego position while keeping the ego heading pointed to the right.
Finally, the predictions are evaluated with:
\begin{itemize}
    \item minADE and minFDE are the minimum ADE and FDE over the $K$ predicted trajectories.
    \item BrierFDE is minFDE with a penalty term $(1-p)^2$, where $p$ is the probability of the best predicted trajectory. This metric encourages the model to assign high confidence to accurate predictions.
    \item Miss Rate is the percentage of samples where minFDE exceeds a certain threshold (e.g., 2 m).

\end{itemize}

To summarize, Table~\ref{tab:task-summary} compares the input, output, and evaluation of the two tasks on SongdoDrive.

\begin{table*}[ht]
\centering
\caption{Summary of the trajectory planning and motion prediction tasks.}
\label{tab:task-summary}
\resizebox{\textwidth}{!}{%
\begin{tabular}{@{}lllllll@{}}
\toprule
Task & Input data & \makecell[l]{Input\\horizon} & Output data & \makecell[l]{Output\\horizon} & Frequency & Metrics \\
\midrule
Trajectory planning & \makecell[l]{Rendered multi-camera semantic views,\\ ego motion history, high-level direction} & 2s & Ego trajectory & 4s & 2Hz & ADE, FDE, TTC infraction rate, NCT\\
\midrule
Motion prediction & \makecell[l]{Map structures and ego/surrounding-\\vehicle trajectories} & 2s & \makecell[l]{6 ego trajectories\\and probabilities} & 6s & 10Hz & minADE, minFDE, Miss Rate, BrierFDE \\
\bottomrule
\end{tabular}%
}
\end{table*}

\section{Experiments}
\label{sec: experiments}

\subsection{Trajectory Planning Settings and Results}
\label{subsec: trajectory planning settings and results}

We adapt two state-of-the-art methods from NavSim \citep{dauner2024navsim} to SongdoDrive, i.e., DrivoR \citep{kirby2026drivor} and RAP \citep{feng2026rap}.
Both models generate multiple trajectory proposals first, and then use a learned scorer to choose the trajectory with the best Predictive Driver Model Score (PDMS).
An important difference is that RAP trains the image encoder to align the distributions of real camera images and the synthetic semantic views.
In this work, we train them to imitate the ground-truth driving trajectories without any other auxiliary scores for simplicity.
For DrivoR, we use the negative of the average of ADE and FDE as the score.
For RAP, we use its alternative option based on Rater Feedback Score \citep{ettinger2021waymodataset}.

\begin{table*}[ht]
    \centering
    \caption{Overall standard-split trajectory planning results.}
    \label{tab:trajectory-planning-main}
    \begin{tabular}{llcccc}
        \toprule
        Method & Training & ADE (m) $\downarrow$ & FDE (m) $\downarrow$ & TTC (\%) $\downarrow$ & NCT (\%) $\downarrow$ \\
        \midrule
        DrivoR & Zero-shot & 3.702 & 8.855 & 38.10 & 8.24 \\
        DrivoR & Full Data & 1.589 & 3.774 & 10.82 & 4.80 \\
        \midrule
        RAP & Zero-shot & 3.399 & 8.116 & 25.56 & 1.12 \\
        RAP & Full Data & \textbf{1.263} & \textbf{2.730} & \textbf{5.89} & \textbf{0.13} \\
        \bottomrule
    \end{tabular}
\end{table*}

Table~\ref{tab:trajectory-planning-main} shows the overall trajectory planning results using the standard train-test splits.
In the zero-shot experiments (Z), the author-released checkpoints of DrivoR and RAP are tested directly on SongdoDrive.
In the full data experiments, DrivoR is trained from scratch for 20 epochs and RAP is fine-tuned from the zero-shot checkpoint for 5 epochs since a full retrain is not possible without paired real and synthetic images.
Both models are trained with a batch size of 32 on two H100 GPUs under a similar budget, where DrivoR requires $\sim$3 hours/epoch and RAP requires $\sim$10 hours/epoch.

When directly applied outside their training domain, both models have unsatisfactory performance.
The high ADE, FDE and TTC infraction rate show that the planners deviate from reasonable human choices and frequently pose risks.
Although DrivoR's Navsim performance is very close to RAP's (93.1 vs 93.7 PDMS), its zero-shot performance on SongdoDrive is less favorable than RAP's, and this difference results from the image appearance gap.
Since RAP was trained to align the rendered semantic views with real camera images, the rendered semantic views are closer to its training domain.
In contrast, DrivoR was trained only with real images, and its domain gap is larger than RAP's.
The NCT rate provides a stronger indication.
As shown in Figure~\ref{fig: rendering example}, the semantic view highlights the road boundary in red, and RAP's regulatory adherence can be better preserved, resulting in a much lower off-road rate of 1.12\%.

Compared to the zero-shot experiments, both trained models improve significantly on all metrics, showing the effect of in-domain supervision.
The overall FDEs of RAP and DrivoR have been reduced by 66.4\% and 57.4\%, respectively.
The fact that RAP improves more suggests that image-domain alignment can facilitate the learning of city-specific driving behaviors.
More exciting progress is observed in safety and regulatory adherence, although they are not explicitly required in training.
The TTC infraction rate of DrivoR drops from 38.1\% to 10.82\% and the metric for RAP decreases from 25.56\% to 5.89\%, which means both models have learned to navigate more safely and RAP does even better with its aligned image domain knowledge.

\begin{figure}[ht]
    \centering
    \includegraphics[width=\columnwidth]{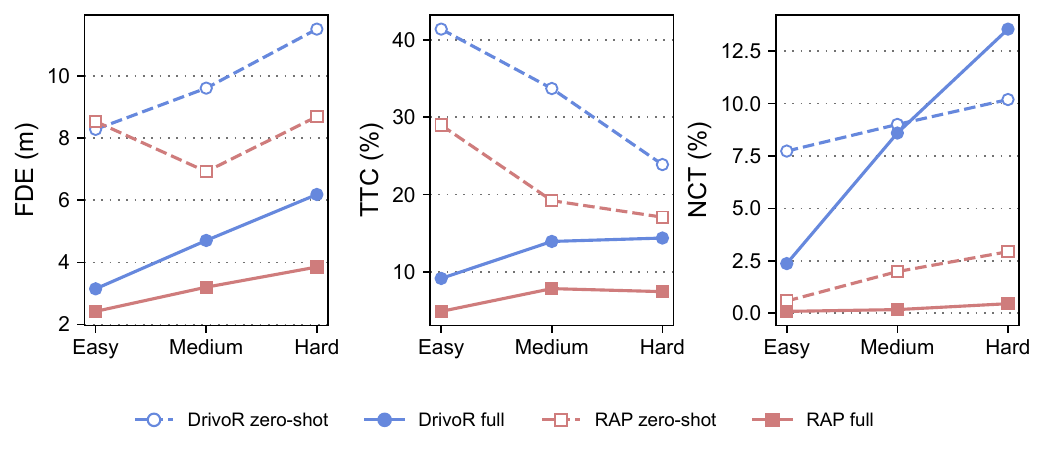}
    \caption{Evaluation results by Kalman difficulty. Easy $\in$ [0, 10), medium $\in$ [10, 20), hard $\geq$ 20.}
    \label{fig: kalman difficulty results}
\end{figure}

\begin{figure}[ht]
    \centering
    \includegraphics[width=\columnwidth]{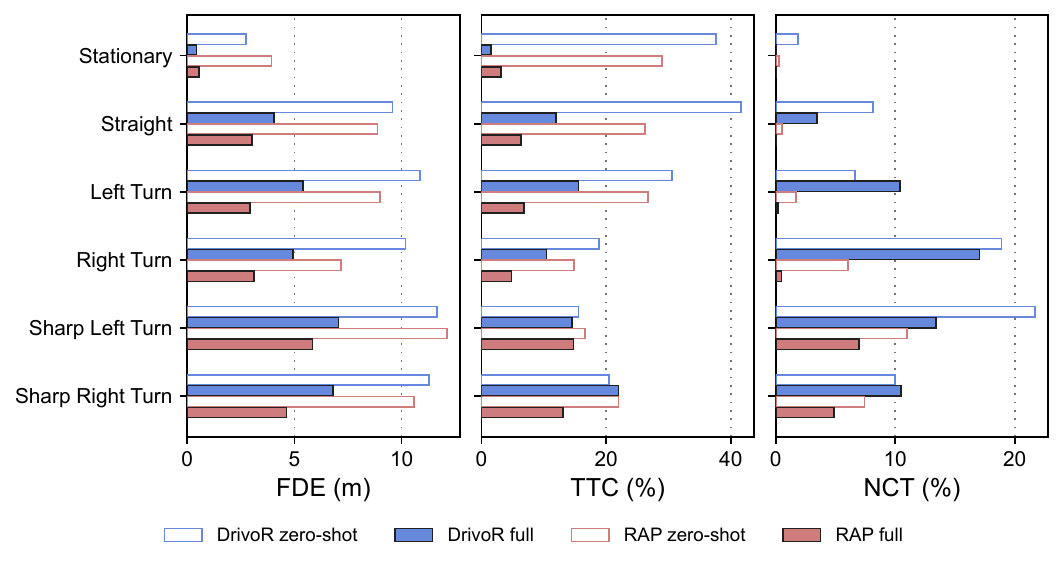}
    \caption{Trajectory planning results by trajectory type}
    \label{fig: trajectory type results}
\end{figure}

To better analyze the performance under different driving scenarios, Figure~\ref{fig: kalman difficulty results} and Figure~\ref{fig: trajectory type results} show the FDE, TTC and NCT metrics of the four tested models by Kalman Difficulty and trajectory types respectively.
Generally, both the trained RAP and DrivoR have better performance in easier scenarios, e.g., the easy category by Kalman Difficulty and the stationary or straight categories by trajectory type.
In more challenging scenarios, e.g., sharp turns, the trained models have less favorable performance.
In contrast, a clear failure mode for zero-shot methods occurs in the easy cases, e.g., high FDE and TTC on stationary and straight trajectories.
This difference suggests that the planners are confused about moving and stopping, and after supervised training with drone data, this distinction can be learned much better.

\subsection{Data Scaling and Cross-Intersection Performance}
\label{subsec: data scaling and cross intersection performance}

This section further investigates how much data is needed to have a significant performance gain.
To make the data scaling compatible with feasible real-world drone monitoring, we make subsets of the standard training set by monitoring sessions identified by intersection, day and time slot.
The sessions are selected via round-robin scheduling over intersections to spread the monitoring efforts evenly over the space.
For example, the 1\% subset has 6 sessions out of the 680 training sessions, and they belong to 6 random intersections.
The model is trained with the same hyperparameters on the training subset and evaluated on the complete test set.

\begin{figure}[ht]
    \centering
    \includegraphics[width=\columnwidth]{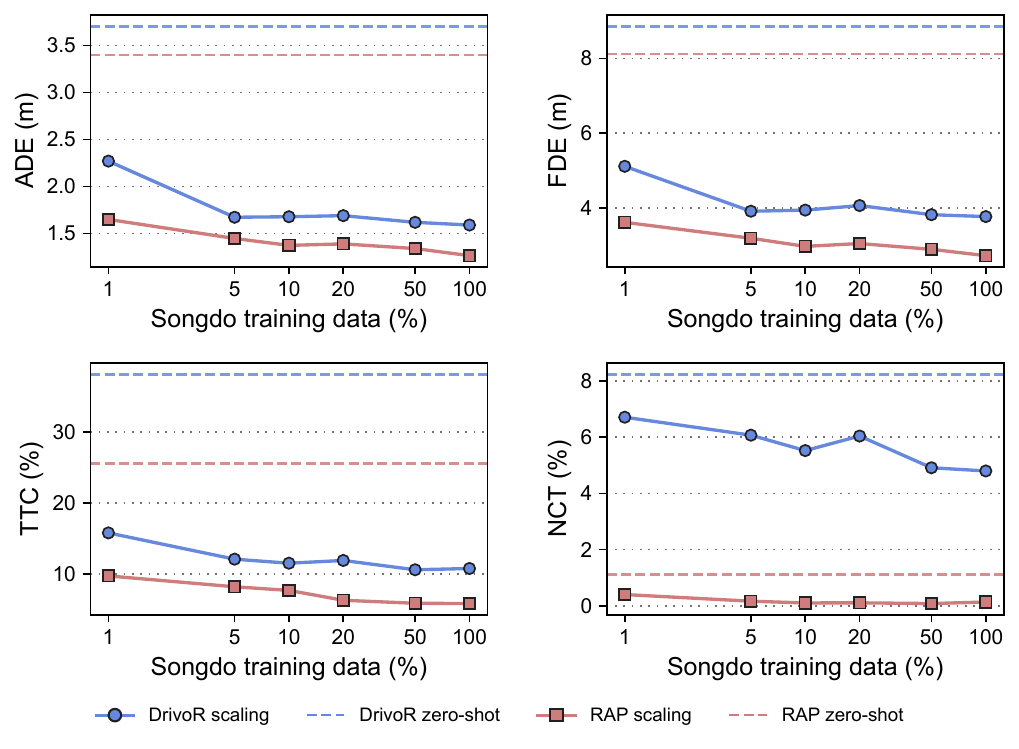}
    \caption{Scaling experiment results. The experiments with 1\%, 5\%, 10\%, 20\% and 50\% training data have 6, 34, 68, 136 and 340 monitoring sessions respectively, corresponding to 1.0, 5.8, 11.7, 23.3 and 58.3 hours of approximate traffic monitoring time.}
    \label{fig: scaling results}
\end{figure}

Figure~\ref{fig: scaling results} shows the data scaling curve for the four metrics with the zero-shot performance as a reference.
The results show that even 1\% data supervision can make a significant difference, with all metrics substantially improved.
While RAP consistently improves with more training data, DrivoR's performance improves much more slowly, which highlights the importance of aligning the image representation space.

In addition to domain gaps across cities and shifts in image appearance, generalization gaps may also exist among different locations in the same city.
To provide a quantified analysis, we make a cross-intersection data split, train the models on 17 intersections and test on the remaining three. 
Meanwhile, the standard train split covers all intersections.



To make a fair comparison, we take the overlapping samples from the test sets of the standard (S) and cross-intersection (C) splits, and evaluate the models trained with the corresponding train sets.
Table~\ref{tab:trajectory-planning-cross-intersection-overlap} shows those evaluation results, and again confirms the similarity of different intersections in the same region.
The models trained with the standard split are generally preferable, since they are trained with data from all locations, but the cross-intersection gaps are not significant.
The only relatively large performance gap is observed for the NCT of DrivoR, where the cross-intersection model struggles more with the road boundaries.

\begin{table}[ht]
    \centering
    \caption{Matched standard (S) and cross-intersection (C) results on overlapping samples}
    \label{tab:trajectory-planning-cross-intersection-overlap}
    \resizebox{\columnwidth}{!}{%
    \begin{tabular}{@{}lcccc@{}}
        \toprule
        Method & ADE (m) $\downarrow$ & FDE (m) $\downarrow$ & TTC (\%) $\downarrow$ & NCT (\%) $\downarrow$ \\
        \midrule
        DrivoR (S) & \textbf{1.485} & \textbf{3.514} & 8.44 & \textbf{4.72} \\
        DrivoR (C) & 1.535 & 3.659 & \textbf{7.94} & 7.87 \\
        \midrule
        RAP (S) & \textbf{1.145} & \textbf{2.447} & \textbf{4.07} & 0.09 \\
        RAP (C) & 1.170 & 2.506 & 5.09 & \textbf{0.06} \\
        \bottomrule
    \end{tabular}
    }
\end{table}

\subsection{Motion Prediction Settings and Results}
\label{subsec: motion prediction results}
In this section, we benchmark the performance of AutoBot~\citep{girgis2022latent}, MTR~\citep{shi2022mtr}, and Wayformer~\citep{nayakanti2023wayformer}.
The implementation is based on UniTraj and default configurations are kept for all methods.
Table~\ref{tab:evaluation-results} presents the overall evaluation metrics.
MTR* denotes an MTR model trained on nuScenes and evaluated zero-shot on SongdoDrive~\citep{caesar2020nuscenes}.

\begin{table}[ht]
    \centering
    \caption{Motion prediction results. Best values in bold.}
    \label{tab:evaluation-results}
    \resizebox{\columnwidth}{!}{%
    \begin{tabular}{lcccc}
        \toprule
        Method & BrierFDE & minADE & minFDE & Miss Rate \\
        \midrule
        MTR* & 3.953 & 1.497 & 3.369 & 0.494 \\
        \midrule
        AutoBot & 1.847 & 0.588 & 1.197 & 0.154 \\
        Wayformer & \textbf{1.763} & \textbf{0.564} & \textbf{1.157} & \textbf{0.148} \\
        MTR & 1.902 & 0.680 & 1.516 & 0.273 \\
        \bottomrule
    \end{tabular}
    }
\end{table}

\begin{table}[ht]
    \centering
    \caption{BrierFDE by Kalman difficulty. Best values in bold.}
    \label{tab:brierfde-kalman-difficulty}
    \resizebox{\columnwidth}{!}{%
    \begin{tabular}{@{}lccc@{}}
        \toprule
        Method & Easy [0, 30) & Medium [30, 50) & Hard [50, 100] \\
        \midrule
        MTR* & 3.248 & 8.088 & 16.975 \\
        \midrule
        AutoBot & 1.722 & 2.600 & 3.564 \\
        Wayformer & \textbf{1.645} & \textbf{2.482} & \textbf{3.242} \\
        MTR & 1.765 & 2.716 & 4.107 \\
        \bottomrule
    \end{tabular}
    }
\end{table}

\begin{table}[ht]
    \centering
    \caption{BrierFDE by trajectory type. Best values in bold.}
    \label{tab:brierfde-trajectory-type}
    \resizebox{\columnwidth}{!}{%
    \begin{tabular}{@{}lcccc@{}}
        \toprule
        Type / Method & MTR* & AutoBot & Wayformer & MTR \\
        \midrule
        Stationary     & 2.017  & 0.623          & 0.562          & \textbf{0.457} \\
        Straight       & 2.704  & 1.791          & \textbf{1.690} & 1.817 \\
        Straight-right & 5.337  & 2.700          & \textbf{2.631} & 2.985 \\
        Straight-left  & 4.549  & 2.423          & \textbf{2.262} & 2.668 \\
        Right turn     & 7.825  & 2.459          & \textbf{2.399} & 2.667 \\
        Left turn      & 7.131  & 2.396          & \textbf{2.350} & 2.554 \\
        Right U-turn   & 6.546  & 6.222          & \textbf{4.857} & 5.993 \\
        Left U-turn    & 12.252 & \textbf{4.513} & 4.951          & 5.024 \\
        \bottomrule
    \end{tabular}%
    }
\end{table}


Among the models trained on SongdoDrive, Wayformer achieves the best performance across all metrics, while AutoBot is a close second.
Notably, the MTR model is a larger model with higher per-epoch training time, but it still delivers slightly worse performance than the lightweight AutoBot.
Compared to the trained MTR model, the zero-shot MTR* has 108\% higher BrierFDE and 120\% higher minADE.
This severe degradation confirms the pronounced domain shift and the necessity of in-domain supervision.

Table~\ref{tab:brierfde-kalman-difficulty} shows the BrierFDE by Kalman difficulty, where the errors of all methods increase from easy to hard cases.
Wayformer remains the best method across all three difficulty levels, with a more obvious advantage on the hard subset.
Meanwhile, compared to zero-shot inference, the trained MTR model achieves a 75.8\% improvement on the hard cases.
Table~\ref{tab:brierfde-trajectory-type} shows a per-type breakdown.
Consistent with the stop-and-go confusion in Figure~\ref{fig: trajectory type results}, MTR* has the largest relative increase in error (+341\% compared to trained MTR) for the stationary type.
Other than that, the straight type is easier to handle, whereas turnings are more challenging.

\section{Conclusions}
\label{sec: conclusions}
In this work, we introduce \emph{SkyDrive} to turn drone-based traffic monitoring data into a supervision source for autonomous driving.
Owing to the extended aerial field of view, \emph{SkyDrive} can efficiently obtain diverse and high-quality trajectories from many vehicles, and create more training samples within the same operation time.
We quantified the generalization gap when applying trajectory planners and motion predictors directly in an unseen city, and identified confusion between stopping and moving as a common failure mode.
The experiments with RAP and DrivoR demonstrated that trajectory planners can be significantly improved with a small amount of data corresponding to $\sim$30 minutes of traffic monitoring per location, and also highlighted the importance of training the image encoder to understand the semantic layout of driving scenarios.
Our investigation has shown that drone-based traffic monitoring is a favorable choice for adapting autonomous driving models to a new city, and we hope this finding can offer a new perspective for practitioners in the autonomous driving community.

\bibliography{aaai2027}


\end{document}